\documentclass[sigconf]{acmart}
\acmISBN{}
\acmDOI{}
\acmConference[ICEA2024]{2024 International Conference on Intelligent Computing and its Emerging Applications}{November 28 – 29, 2024}{Tokyo, Japan}

\usepackage{caption}
\usepackage{subcaption}
\usepackage{bbding}

\AtBeginDocument{%
  }

\begin{document}

\title{LiDAR-Camera Fusion for Video Panoptic Segmentation without Video Training}

\author{Fardin Ayar$^1$, Ehsan Javanmardi$^2$, Manabu Tsukada$^2$, Mahdi Javanmardi$^1$, Mohammad Rahmati$^1$}
\email{fardin.ayar@aut.ac.ir, ejavanmardi@g.ecc.u-tokyo.ac.jp, tsukada@hongo.wide.ad.jp, mjavan@aut.ac.ir, rahmati@aut.ac.ir}
\affiliation{%
\institution{$^1$Department of Computer Engineering, Amirkabir University of Technology, Tehran, Iran \\
$^2$Graduate School of Information Science and Technology, The University of Tokyo, Tokyo, Japan}
\country{}}


\begin{abstract}
Panoptic segmentation, which combines instance and semantic segmentation, has gained a lot of attention in autonomous vehicles, due to its comprehensive representation of the scene.
This task can be applied for cameras and LiDAR sensors, but there has been a limited focus on combining both sensors to enhance image panoptic segmentation (PS). Although previous research has acknowledged the benefit of 3D data on camera-based scene perception, no specific study has explored the influence of 3D data on image and video panoptic segmentation (VPS).
This work seeks to introduce a feature fusion module that enhances PS and VPS by fusing LiDAR and image data for autonomous vehicles. We also illustrate that, in addition to this fusion, our proposed model, which utilizes two simple modifications, can further deliver even more high-quality VPS without being trained on video data. The results demonstrate a substantial improvement in both the image and video panoptic segmentation evaluation metrics by up to 5 points.
\end{abstract}



\maketitle

\section{Introduction}
Multimodal-learning-based sensor fusion has consistently been one of the potential solutions for enhancing the performance of computer vision models. In multimodal learning, the input to the machine learning algorithms consists of more than one type of data. This approach can lead to significant improvements, especially in cases where there is a high degree of redundancy between input modalities \cite{wang2019multi}. One prominent example is the fusion of image and 3D data\footnote{The way we treat LiDAR data is by projecting it onto the 2D image plane and up-sampling it to become dense. Therefore, there is no difference in our proposed model in terms of what the 3D modality is.}
(e.g. from LiDAR or depth images), which has been extensively studied and proven to positively affect detection-based algorithms \cite{caltagirone2019lidar, geng2020deep}. 

\begin{figure}[t]
    \centering
    \begin{subfigure}[b]{0.49\columnwidth}  
      \centering
      \includegraphics[width=\textwidth]{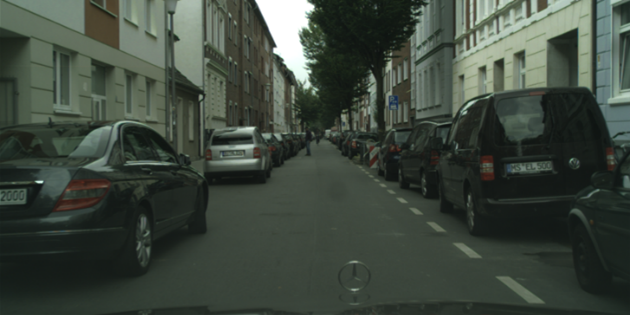}
      \caption{Example image}
      \label{fig:grid_gaussian_sub-first}
    \end{subfigure}
    \hfill
    \begin{subfigure}[b]{0.49\columnwidth}  
      \centering
      \includegraphics[width=\textwidth]{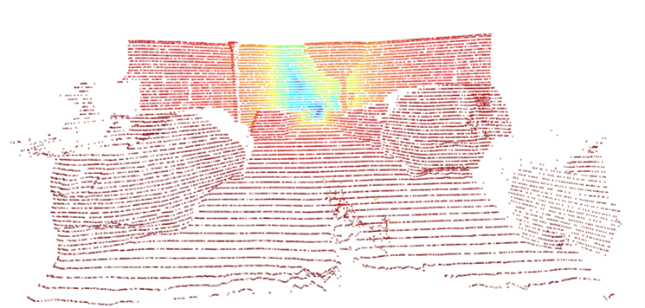}
      \caption{Simulated semi-sparse LiDAR}
      \label{lidar}
    \end{subfigure}
    \vskip\baselineskip 
    \begin{subfigure}[b]{0.49\columnwidth}  
      \centering
      \includegraphics[width=\textwidth]{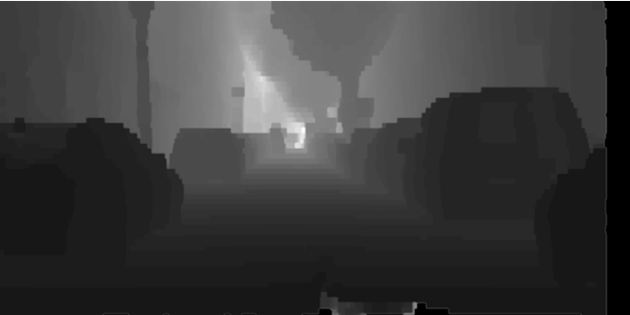}
      \caption{Up-sampled LiDAR depth}
      \label{fig:grid_gaussian_sub-third}
    \end{subfigure}
    \hfill
    \begin{subfigure}[b]{0.49\columnwidth}  
      \centering
      \includegraphics[width=\textwidth]{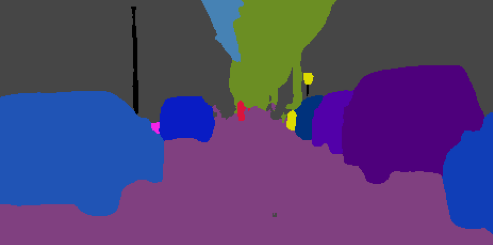}
      \caption{Panoptic segmentation}
      \label{fig:panoptic}
    \end{subfigure}
    \caption{(a) Example image from the cityscapes dataset. (b) Semi-sparse LiDAR points simulated from estimated disparity \cite{xu2023unifying}. (c) Upsampled LiDAR points using \cite{ku2018defense}. (d) Output of the panoptic segmentation model.}
    \label{fig:visualization}
\end{figure}

Although there has always been research on the fusion of image-LiDAR in the context of AVs, this fusion has generally been done to improve LiDAR segmentation \cite{wang2019multi} tasks due to the lack of contextual clues in point clouds. Specifically, few works have explored the potential improvements from this fusion in the context of panoptic image segmentation (Figure \ref{fig:panoptic}) \cite{sodano2023robust}, which is the task of segmenting things (e.g. cars, people, etc.) and segmenting (e.g. sky, road, etc.) simultaneously in images \cite{kirillov2019panoptic} or videos \cite{kim2020video}.

 Previous works have demonstrated that transformer-based segmentation methods, such as Mask2Former, can be used for online video tasks without video training \cite{huang2022minvis}. These studies have shown that object queries in transformer-based methods offer highly discriminative representations of objects in a scene, allowing for the effective matching of objects across consecutive frames without requiring video training losses like contrastive loss \cite{idol}.

 According to these, our aim is to improve both image panoptic segmentation (PS) and video panoptic segmentation (VPS). Firstly, we propose a technique that focuses on fusing features extracted from two separate ResNet50 networks for image and LiDAR data, regardless of the specific panoptic segmentation model. We will demonstrate that implementing this method within a Mask2Former network \cite{cheng2022masked} can improve performance based on evaluation metrics.

 In the second part of this work, we will demonstrate that a video panoptic segmentation model without video training for autonomous vehicle datasets does not yield competitive results compared to models trained on video data. To address this, in addition to the above-mentioned sensor fusion method, we propose two modifications to the network. These changes, which do not add computational cost or require video training, aim to minimize the gap between video-trained and video-free models. The first change involves adding spatial information to the output queries via an additional loss function. The second involves reusing the queries from the previous frame in the current frame. As will be shown in our experiments, these changes provide a significant improvement over the baseline model. 
 
 Overall, Our contributions are as follows:
\begin{itemize} 
 \item Proposed a method for the fusion of LiDAR and image data using a novel feature fusion technique, enhancing panoptic segmentation performance.
 \item Introduced location-aware and time-aware queries to improve object tracking and reduce mismatching in video panoptic segmentation.
 \item Achieved significant improvements in video segmentation quality without video-specific training, demonstrating the potential for autonomous vehicle applications.
\end{itemize} 

\section{Related Work}
\subsection{Panoptic Segmentation}
Transformer-based panoptic segmentation methods \cite{cheng2022masked} have gained popularity over multi-task approaches \cite{cheng2020panoptic} in recent works, inspired by the seminal work of Carion et al. \cite{carion2020end}. In contrast to Carion et al., who used transformer queries to extract bounding boxes for each thing/stuff, Wang et al. \cite{wang2021max} treated queries as convolutional filters. By convolving them with feature maps, they extracted masks for each thing/stuff in the input images. Cheng et al.'s MaskFormer \cite{cheng2021per} introduced a unified framework for all three segmentation tasks (instance, semantic, and panoptic) that outperformed traditional pixel-wise segmentation. Later, there were extensions of this idea to improve efficiency and/or efficacy, one of which was the pioneering work of Mask2Former \cite{cheng2022masked}. Mask2Former uses masked attention to limit query attention to predicted segmentation areas, significantly improving performance across segmentation tasks. Since then, research in this area, in collaboration with recent advances in transformer-based detection architectures, has continued to improve the efficiency of panoptic segmentation \cite{li2023mask}.

\subsection{Online Video Panoptic/Instance Segmentation}
Given the inherent similarity between the problems of panoptic segmentation and instance segmentation, as well as the irrelevance of stuff tracking, most methods presented for the video versions of both tasks are shared. Hence, we do not distinguish between video instance segmentation and video panoptic segmentation \cite{kim2020video} tasks.

Following the transformer-based models in segmentation, recent works in online video segmentation extended these models to track objects between consecutive frames. The majority of works used transformer queries as a representative of the corresponding object and track objects by matching queries between the two frames \cite{idol}. To make these queries more discriminative researchers usually use methods like contrastive loss, memory bank, or other ways to update queries \cite{li2023tcovis}.

Hong et al. \cite{huang2022minvis} demonstrated that even without training on video datasets and using only a bipartite matching algorithm between queries in two consecutive frames, competitive performance can be achieved compared to video-trained models. Lee et al. \cite{li2022video} examined this observation for video panoptic segmentation but did not achieve competitive results, likely due to the more complicated nature of video panoptic segmentation. In this paper, we aim to use a similar method to develop a \textit{video-free} video panoptic segmentation model. We intend to keep the main idea of not using video training and propose simple ideas to improve the baseline approach of Hong et al \cite{huang2022minvis}.

\subsection{LiDAR-Image Multimodal Learning}
Fusing 3D data (stereo depth or LiDAR) with images has a long history in object detection and semantic segmentation. Early work by Gupta et al. \cite{gupta2014learning} demonstrated that adding depth information to camera images improves tasks such as region proposal, object detection, semantic, and instance segmentation, likely due to the strong edge information in the depth images. 
The following research in this area has primarily focused on semantic segmentation \cite{zhang2023cmx}. For panoptic segmentation, specifically, there are few studies \cite{sodano2023robust}, \cite{fischedick2023efficient} that address the combination of depth and image data. These studies have evaluated their proposed networks on internal datasets with the assumption of having dense depth maps.

To date, no research has addressed panoptic segmentation using a combination of image and lidar data on autonomous vehicle datasets, although there are works that fused image and dense depth data for PS. \cite{sodano2023robust} utilized a dual backbone framework to extract features from both depth and image data, and then fused them using an adaptive weighting scheme. Although most recent studies have adopted a similar dual-backbone framework, Fischedick et al. \cite{fischedick2023efficient} suggested that a single unified transformer could potentially replace the dual-backbone architecture.
The closest work to this research is by Gang et al.  \cite{geng2020deep}, who used lidar and camera data to improve object segmentation. Their experiments showed that lidar data could enhance the performance of object detection models.
The current study aims to fuse lidar data into an existing camera-based panoptic segmentation framework to improve it in both image and video domains. 

\section{Proposed Method}
\begin{figure*}[h]
  \centering
  \includegraphics[width=\linewidth]{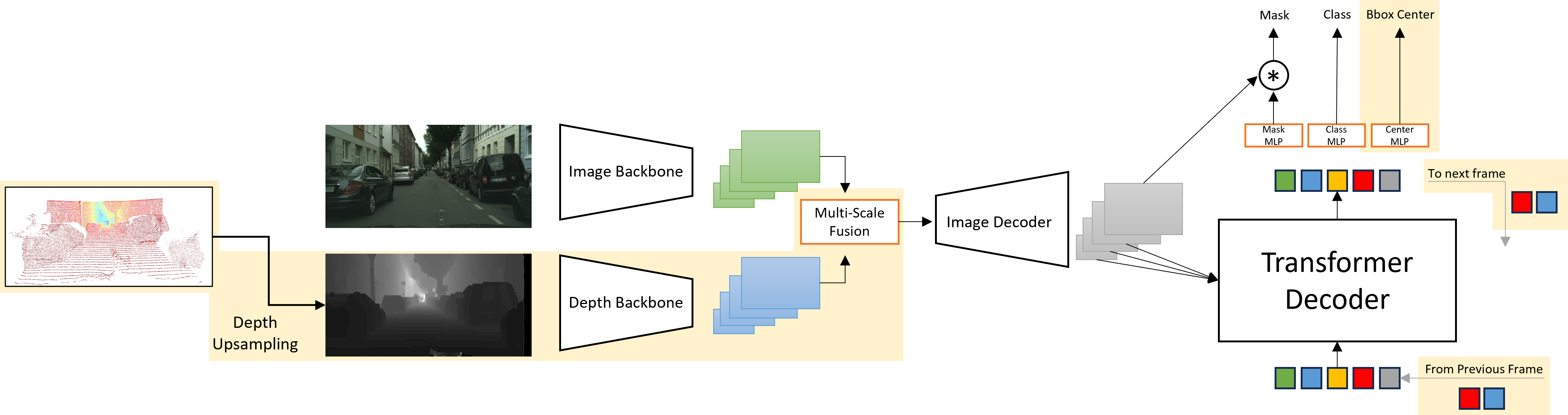}
  \caption{Overall architecture of our proposed method, which is based on Mask2Former \cite{cheng2022masked}. The parts with a yellow background are our contributions}
 \label{mask2form}
\end{figure*}

This section introduces the proposed method of this paper. We begin with a concise overview of the base network utilized in this study. Next, we discuss the proposed improvements in two parts: first, we explore enhancements to the base panoptic segmentation network using depth data obtained from stereo and LiDAR. Second, we transfer these improvements to the video domain, aiming to propose more specialized ideas for video panoptic segmentation within the base framework. The overall architecture of our model is provided in Figure \ref{mask2form}.
\subsection{Base Model}
As mentioned in the previous section, the base network used in this paper is Mask2Former \cite{cheng2022masked}. Given an image \( I \in \mathbb{R}^{3 \times H \times W} \), a backbone network (such as ResNet) extracts multi-scale features. These features are fed into a pixel decoder, which produces multiscale feature maps. 

A 9-layer transformer decoder network starts with \( N \) fixed initial queries \( X_0 \in \mathbb{R}^{C_x \times N} \) (Also known as object queries) as the target sequence and receives feature maps as the source sequence in each layer. It updates the initial queries using a masked cross-attention mechanism and a self-attention mechanism.

Output queries \( X_L \in \mathbb{R}^{C_x \times N} \) will further pass through two separate MLPs to produce class probabilities and mask filters that are convolved with the largest pixel decoder's feature maps to produce output masks for both things and stuff.

\subsubsection{Extension to Video Panoptic Segmentation}\label{etension}

As mentioned in Chapter 2, we follow Huang et al. \cite{huang2022minvis} to extend the base model to video. After extracting the object queries of the current frame \( X_L^t \) and the object queries of the previous frame \( X_L^{(t-1)} \), the matching between segments of two consecutive frames is performed using the Hungarian algorithm. In this case, not only is there no need for a separate module to learn segment matching (as in \cite{kim2020video} or other works), but the network also does not require video datasets for training.

\subsection{Improving Image Panoptic Segmentation Using Depth}

In the first part of this work, we explore the use of depth obtained from stereo vision and LiDAR as an auxiliary input in panoptic segmentation. The aim of this section is to propose a method for fusing image and depth data that is independent of network architecture and can be applied to the backbone network. The following goals and motivations are considered in the design:

\begin{itemize}
    \item[(a)] Since the backbone network is a crucial and shared component of all networks in detection-based tasks, it can be enhanced with additional depth input, regardless of the downstream task or the specific network architecture.
    \item[(b)]LiDAR and stereo depth data can be transformed into each other with camera parameters available. Despite LiDAR data being sparser, it is expected that after resampling, LiDAR data can perform similarly or even better than stereo depth. Therefore, our design aims not to make assumptions about the type of input data (LiDAR or stereo depth).
    \item[(c)] Given the depth (z) and the pixel coordinates (u,v), the other two 3D coordinates of each pixel (x,y) will be a factor of \( (u.z,v.z) \). Thus, unlike some other studies, we assume that within the image boundaries, using depth as the only additional input will be sufficient.
\end{itemize}

Based on the above explanations, the framework for the proposed method of fusing image and depth data is introduced in the following sections. Given a three-channel camera image \( I \in \mathbb{R}^{3 \times H \times W} \) and a single-channel depth image \( D \in \mathbb{R}^{1 \times H \times W} \), we can use two entirely separate networks to extract multi-scale features from the image \( F_l^I \in \mathbb{R}^{C_l^I \times H_l \times W_l} \) and from the depth \( F_l^D \in \mathbb{R}^{C_l^D \times H_l \times W_l} \) while \(l \in \{1,2,3,4\}\) being index of multiscale features from the backbone. It is important to note that while the spatial dimensions of the multi-scale features from both networks are the same, their channel dimensions can be different. The enhanced multi-scale features \( F_l^E \in \mathbb{R}^{C_l \times H_l \times W_l} \) are obtained using a combination of these two.

\[
F_l^E = \phi(F_l^I,F_l^D) \tag{3-1}
\]

In the above relation, \( \phi \) is the feature combination function, which can range from simple addition to more complex functions such as attention mechanisms. Subsequently, the combined features can be fed into downstream networks, such as Mask2Former. In the experiments, ResNet50 is used as the backbone for both networks (\( C_l^I=C_l^D=C_l\)); however, this formulation can also be applied to other feature extraction networks. Additionally, the following function has been used for \( \phi \):

    \[
    \phi(F_l^I,F_l^D) = F_l^I + \sigma(\text{conv}_{1 \times 1}(F_l^I)) \cdot \gamma F_l^D \tag{3-2}
    \]
    According to the above equation, our idea is that the decision on the extent to which depth features are incorporated at each spatial location is dynamically made based on the image features at the same location.
    where \( \gamma \in \mathbb{R}^{C^D} \) is learned during training.

\subsection{Improving Video Panoptic Segmentation}

Applying a video-free approach like Huang et al. \cite{huang2022minvis} for video panoptic segmentation, particularly for datasets related to autonomous vehicles where the number of objects and the similarity between them may be high, will likely lead to a reduction in accuracy compared to networks that use video supervision. However, the absence of a separate module for video learning reduces the network's complexity and computational load. Additionally, this method can demonstrate the extent to which the use of depth in feature extraction influences the ability to obtain discriminative queries.

As we will see, the proposed sensor fusion alone cannot yield competitive results when compared to video panoptic segmentation networks that are trained with video supervision. Thus, we aim to add simple modifications to the transformer's queries to improve video panoptic segmentation while keeping the model video-free.
 
\subsubsection{Location-Aware Queries (LAQ)}

The current method of using queries for matching segments between frames is heavily reliant on object appearance, which can lead to errors. To address this, we propose a mechanism to estimate the 2D position of each segment using a three-layer MLP network. This approach makes the queries position-aware, reducing errors when matching segments with similar appearances. Additionally, it eliminates the need for manual parameter tuning or heuristic methods. The network is trained to predict the coordinates of the center of the bounding box for each segment in a supervised manner, using an L1 loss function. Given the shapeless and often fragmented nature of stuff (e.g. sidewalks), predicting positions for them with this method seems meaningless; thus, this loss function is only calculated for object segments. 

\subsubsection{Time-Aware Queries (TAQ)}

The initial Mask2Former queries \(X_0\) serve as initial region proposals that become more refined throughout the transformer decoder. This concept could be applied in the video domain to improve network performance. Output queries from the previous frame could be used as raw input queries for the transformer decoder in the current frame, leveraging the segmented regions from the previous frame as region proposals for the current frame. This modification does not change the fundamental nature of the proposed method and will only be applied during the evaluation phase while network training remains unchanged.

Based on our experiments, we found that using all queries from the previous frame for the current frame is not the best approach. Queries with empty classes do not provide significant information about the previous frame. Reusing them might divert the model's attention from identifying new objects in the current frame. As a result, only non-empty queries from the previous frame are utilized in the current frame.

Note that the concept of TAQ has been utilized in recent studies under different names, such as in \cite{li2023tcovis, idol}. However, in all of these works, TAQ is employed during training to enable the model to learn to reuse the queries. Our experiments have demonstrated that incorporating TAQ into the video-free video panoptic segmentation model does not lead to improved results. This is likely due to queries being assigned to different objects in the subsequent frame, resulting in mismatching. This mismatching problem was solved with the previous idea of LAQ. Hence, as we will see, This idea actually improves our video-free model.

\section{EXPERIMENT AND EVALUATION}
In this section, we experiment with the proposed methods and analyze their results. We will start with a brief overview of the datasets used for training the networks. Following that, we will outline the settings related to network training. The majority of this section consists of experimental results and their corresponding analysis.
\subsection{Datasets}
We have used the cityscapes dataset \cite{cordts2016cityscapes} and the cityscapes-vps \cite{kim2020video} dataset for evaluating our model on PS and VPS, respectively.  The primary challenge of using these datasets is the absence of LiDAR data. Therefore, we employed a learning-based stereo-matching technique \cite{xu2023unifying} to generate high-quality depth maps. Subsequently, we applied angle-based downsampling to mimic a velodyne64 LiDAR sensor (Figure \ref{lidar}). The simulated LiDAR data was further randomly downsampled by a factor of 0.3 to replicate the ray-drop effect observed in real LiDAR data.
\subsection{Network Training and Evaluation Metrics}
In the training of Mask2Former \cite{cheng2022masked}, a batch size of 16 was used, but in our implementation,  due to memory limitations, it is set to 6. Apart from the batch size, no other changes were made to the training process. Additionally, for the L1 loss function of the location-aware queries, a weight of 5 was used. For more details on training and inference of the models, please refer to \cite{cheng2022masked}.

The evaluation metric for panoptic segmentation is called Panoptic Quality (PQ) and was introduced in a paper by Kirillov et al. \cite{kirillov2019panoptic} The PQ metric can be separately defined for "things" (PQ$^{th}$) and "stuff" (PQ$^{st}$). Expanding this metric to Video Panoptic Segmentation, Kim et al. \cite{kim2020video} proposed VPQ$^k$ using a similar definition as PQ, considering a window size of K. Different values of k has been used to evaluate both segmentation and tracking and the mean VPQ across different k values is considered as the final metric.

\subsection{Pretraining the depth networks}
The ResNet50 depth network with around 20 million parameters has not been pre-trained, which may lead to overfitting, especially in the Cityscapes dataset with only 3000 training samples. To explore the potential of depth data, we will investigate the use of a pre-trained network. For this paper, a depth image classification network was trained from scratch using estimated depth \cite{bhat2023zoedepth} from the ImageNet dataset, as no pre-trained network for depth image classification was available at the time.

\subsection{Results for Panoptic Segmentation}
In this study, we treat stereo and LiDAR depth data in the same manner. Since stereo-depth data typically suffers from significant noise, its accuracy is expected to be lower compared to LiDAR data. However, considering the much lower cost of stereo depth compared to using a LiDAR sensor, it seems worthwhile to explore potential improvements. Note that for this, we have used stereo depth provided by the cityscapes dataset, which is from classical image matching methods and has much lower quality than fig. \ref{fig:visualization}c.

\begin{table*}[h]
\centering
\caption{Results of the base model and the proposed models. All experiments except the first are with batch size 6. PQ$^{th}$ and PQ$^{st}$ refer to the panoptic quality of things and stuff, respectively.}
\begin{tabular}{|l|c|c|c|c|c|l|}
\hline
Depth Data Type & PQ & PQ$^{th}$ & PQ$^{st}$ & Combination Method & Description \\ \hline
Without Depth  & 61.10 & - & - & - & Base Model (Batch Size 16) \\ \hline
Without Depth  & 57.18 & 45.64 & 65.58 & - & Base Model \\ \hline
Stereo Depth & 58.20 & 46.87 & 66.44 & Sum & \\ \hline
Stereo Depth & 59.28 & 47.46 & 66.49 & Dynamic Weighting & Ours \\ \hline
LiDAR &  60.96 & 51.50 & 67.85 &  Dynamic Weighting & Ours\\ \hline
LiDAR (Pre-trained Depth) & \textbf{62.12} & \textbf{53.61} & \textbf{68.31} & Dynamic Weighting & Ours Final Model\\ \hline
\end{tabular}
\label{tab:results_eval_metrics_pq}
\end{table*}

For using LIDAR data, as mentioned in the previous chapter, after mapping them to the image, we use a fast sampling method \cite{ku2018defense} to convert them into dense depth images.

The results of various experiments are provided in Table \ref{tab:results_eval_metrics_pq}. One key observation is that the performance of the base model decreases significantly when a batch size of 6 is used. This suggests that we should consider using a larger batch size in future work. To assess the effectiveness of our proposed feature fusion function (Equation 3-2), we initially experimented with a simple summation as a feature fusion function. As evident from the table, our proposed fusion function performs better than the simple summation of features. Furthermore, the table indicates that the use of LiDAR data, particularly when the LiDAR backbone is pretrained, can significantly enhance the base model. One notable observation here is that incorporating the depth data improves the segmentation quality of things more than stuff. This is expected as we know that stuff classes are usually defined by their texture rather than their geometry.

\begin{table*}
\centering
\caption{Comparison of the base model with our proposed methods and a video-supervised network.}
\begin{tabular}{|l|l|c|c|c|c|c|c|c|c|c|}
\hline
Model Name & Description & VPQ & VPQ$^{th}$ & VPQ$^{st}$ & VPQ$^{0}$ & VPQ$^{0st}$ & VPQ$^{0th}$ & VPQ$^{5th}$ & VPQ$^{10th}$ & VPQ$^{15th}$ \\ \hline
Base Model & & 51.71 & 30.05 & 67.45 & 60.34 & 70.46 & 46.43 & 30.02 & 23.86 & 19.9 \\ \hline
+ Depth & & 54.36 & 34.00 & 69.64 & 64.39 & 72.63 & 53.07 & 33.41 & 27.35 & 22.15 \\ \hline
+ Location-aware Queries & & 55.41 & 35.63 & 69.81 & 65.10 & 73.36 & 53.74 & 36.04 & 28.61 & 24.11 \\ \hline
+ Time-aware Queries & Proposed Model & \textbf{57.24} & 39.86 & \textbf{69.88} & 65.23 & \textbf{73.40} & 53.31 & 40.88 & 34.81 & 30.39 \\ \hline
Video K-Net \cite{li2022video} & Video Supervised & 57.08 & \textbf{45.0} & 66.9 & \textbf{65.6} & 71.5 & \textbf{57.4} & \textbf{43.4} & \textbf{36.5} & \textbf{33.1} \\ \hline
\end{tabular}
\label{tab:results_eval_metrics}
\end{table*}

\subsection{Results for Video Panoptic Segmentation}
 As mentioned earlier, our proposed method for video panoptic segmentation is video-free, and therefore, the models presented in the previous section can be used here without modification. Following the common practice of related works \cite{li2022video}, all models in this section are first trained on the Cityscapes dataset and then fine-tuned on the Cityscapes-vps for an additional 5,000 iterations.

The results of the proposed methods are shown in Table \ref{tab:panoptic_video_results}. Since the video panoptic quality (VPQ) metric is more important for objects, VPQ$^{th}$ is shown for different K values, while VPQ$^{st}$ is only shown for K=0 and the average case.

As can be seen from the table, using Mask2Former without video supervision does not yield very satisfactory results compared to video-supervised methods; however, using LIDAR data without any additional modifications can improve the VPQ metric by up to 3\% over the baseline model. The use of the introduced query ideas increases this metric by about 5.5\% over the baseline model, bringing it closer to video-supervised networks (Video K-Net \cite{li2022video}). Unfortunately, a large part of this improvement is due to the high quality of stuff segmentation, and for thing segmentation, our proposed model still lags significantly behind supervised methods. Nevertheless, our proposed model improves things' video panoptic quality (VPQ$^{th}$), which is more important, by about 10\% over the baseline. The fact that no video supervision is used during training, along with a smaller batch size and shorter training time, shows the high potential of our proposed method.

\begin{figure}[b]
\centering
\includegraphics[width=\linewidth]{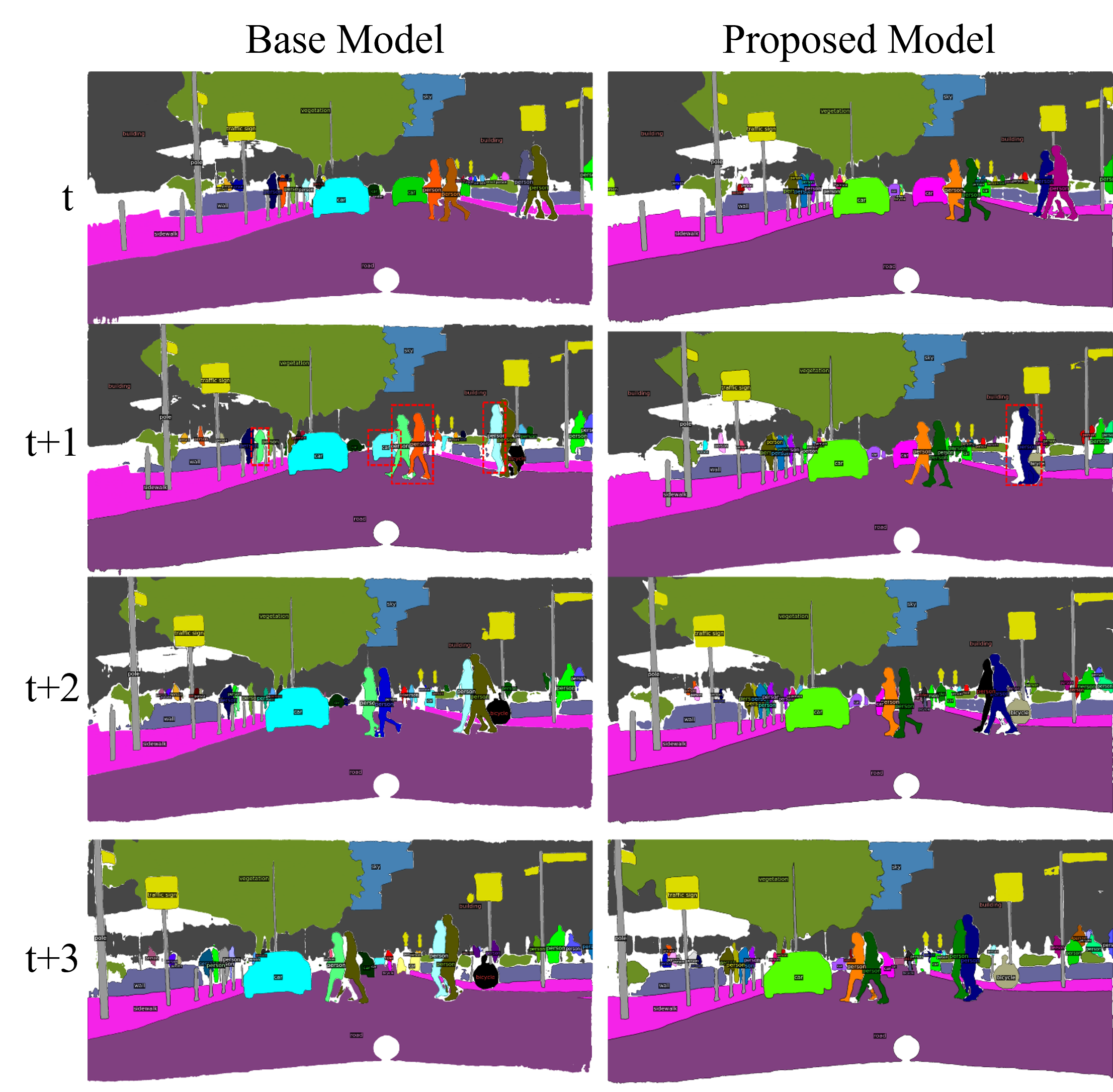}
\caption{Panoptic segmentation output for a video sequence. The base model (left) has significantly more ID switches compared to our proposed method (right). Some ID switches are denoted with bounding boxes.}
\label{fig:video_segmentation}
\end{figure}

In Figure \ref{fig:video_segmentation}, the results of our proposed method and the baseline model are shown for a sample video from the Cityscapes-vps dataset. Each object's color represents its ID. As can be seen, although our proposed method occasionally struggles with matching segments between frames, its errors are significantly fewer compared to the baseline. To emphasize this further, some of the mistakes in the second image of Figure \ref{fig:video_segmentation} are highlighted with a red box.

\section{Conclusion}

This study introduced a novel approach to image/video panoptic segmentation by leveraging LiDAR-camera fusion without the need for video training. The proposed method significantly improves both image and video panoptic segmentation tasks, particularly by incorporating depth information from LiDAR and stereo vision. Through dynamic feature weighting, the model demonstrates notable performance gains over baseline model.

Additionally, the proposed time-aware and location-aware queries enhance video segmentation by mitigating common issues like object mismatching across frames. Although the performance still lags behind video-supervised methods, the improvements seen, particularly in "things" segmentation, highlight the potential of the proposed model for autonomous driving applications.

Overall, the fusion of LiDAR and image data, coupled with the efficient use of transformer-based architectures, opens new avenues for improving panoptic segmentation in scenarios where video training is impractical or unavailable. Further work could involve optimizing batch sizes and exploring more complex fusion techniques to further enhance performance.


\bibliographystyle{ACM-Reference-Format}
\bibliography{sample-base}

\appendix

\end{document}